\pdfoutput=1

\documentclass[11pt]{article}

\usepackage{EMNLP2022}

\usepackage{times}
\usepackage{latexsym}

\usepackage[T1]{fontenc}

\usepackage[utf8]{inputenc}

\usepackage{microtype}

\usepackage{inconsolata}

\usepackage{amssymb}
\usepackage{amsmath}
\usepackage[pdftex]{graphicx}
\usepackage{tabularx}

\usepackage{colortbl}
\definecolor{f05}{rgb}{0.95,0.95,0.95}

\title{Conciseness: An Overlooked Language Task}

\author{Felix Stahlberg \and Aashish Kumar \and Chris Alberti \and Shankar Kumar \\
     Google Research \\ \texttt{\{fstahlberg,kumaraashish,chrisalberti,shankarkumar\}@google.com}}

\begin{document}
\maketitle
\begin{abstract}
We report on novel investigations into training models that make sentences concise. We define the task and show that it is different from related tasks such as summarization and simplification. For evaluation, we release two test sets, consisting of 2000 sentences each, that were annotated by two and five human annotators, respectively. We demonstrate that conciseness is a difficult task for which zero-shot setups with large neural language models often do not perform well. Given the limitations of these approaches, we propose a synthetic data generation method based on round-trip translations. Using this data to either train Transformers from scratch or fine-tune T5 models yields our strongest baselines that can be further improved by fine-tuning on an artificial conciseness dataset that we derived from multi-annotator machine translation test sets. 
\end{abstract}

\section{Introduction}

\begin{quote}
{\em ``Vigorous writing is concise. A sentence should contain no unnecessary words, a paragraph no unnecessary sentences, for the same reason that a drawing should have no unnecessary lines and a machine no unnecessary parts.''}
\begin{flushright}
\citet{elementsofstyle}\\
The Elements of Style
\end{flushright}
\end{quote}

Conciseness is a writing principle of removing redundant information in text. Even though conciseness is highly valued in expository English writing and is often considered good writing style \citep{brock1992teaching,writingwell}, it is still an understudied topic in the natural language processing (NLP) community, mainly due to the lack of annotated data sets. However, automatic methods for improving conciseness have the potential to improve the writing experience even for native speakers, or to provide useful tools for editorial tasks. In this work we take initial steps towards conciseness from an NLP perspective. We release\footnote{\url{https://github.com/google-research-datasets/wiki-conciseness-dataset}} two hand-annotated test sets for conciseness -- {\em Concise-Lite} (2-way annotated) and {\em Concise-Full} (5-way annotated). {\em Concise-Lite} annotators were asked to make minimal changes to the original sentence, whereas {\em Concise-Full} annotators were given the option to make larger rewrites. Table \ref{tab:examples-conciseness} contains examples from both test sets. For evaluation, we compute $F_{0.5}$-scores of edit spans, a metric that is also commonly used for grammatical error correction (GEC) \citep{dahlmeier-ng-2012-better,felice-etal-2016-automatic,bryant-etal-2017-automatic}. Given that both the test sets and the evaluation tool we employ are publicly available, we hope our setup will encourage NLP researchers to investigate models for conciseness.

\begin{table*}[t!]
\centering
\small
\begin{tabularx}{\linewidth}{XXX}
\hline
\textbf{Input sentence} & \textbf{Concise-Lite} & \textbf{Concise-Full} \\
\hline
Gemco had a version called Memco, also owned by Lucky Stores, that operated stores in the Chicago and Washington, D.C., areas. &
Gemco had a version called Memco, \textbf{owned} by Lucky Stores, \textbf{operating} stores in the Chicago and Washington, D.C. & \textbf{Memco was} a version of Gemco \textbf{operated by} Lucky Stores in Chicago and Washington, D.C. \\
\hline
The film was adapted from a best-selling biography of the brothers, and was well presented and well received. & The film was adapted from a best-selling biography of the brothers, and was well presented \textbf{and received}. & The \textbf{film, adapted} from the \textbf{brothers'} best-selling biography, was well presented \textbf{and received}. \\
\hline
\end{tabularx}
\caption{\label{tab:examples-conciseness} Example sentences from our {\em Concise-Lite} and {\em Concise-Full} test sets.}
\end{table*}

\begin{table*}[t!]
\centering
\small
\begin{tabularx}{\linewidth}{XXX}
\hline
\textbf{Input sentence} & \textbf{Abstractive sentence summarization} & \textbf{Conciseness model output} \\
\hline
Exxon corp.\ and Mobil corp.\ have held discussions about combining their business operations, a person involved in the talks said Wednesday. & Exxon and Mobil discuss combining business operations; possible merger. & Exxon Corp.\ and Mobil Corp.\ \textbf{have discussed} combining their business operations, a person involved in the talks said Wednesday. \\
\hline
Chuck Knoblauch and Tino Martinez were as popular as squeegee men a week ago, the speculation rampant that one or the other or both might be exiled if the Yankees' historic year crumbled in the post-season. & Knoblauch and Martinez home run hits cinch Yankee's First World Series game & Chuck Knoblauch and Tino Martinez were as popular as squeegee men a week ago, the speculation rampant that \textbf{either or both could} be exiled if the Yankees' historic year crumbled in the \textbf{postseason}. \\
\hline
\end{tabularx}
\caption{\label{tab:examples-summ-conciseness} Example outputs of one of our conciseness models on sentences from an abstractive sentence summarization data set \citep[DUC2004]{over-duc-summarization}.}
\end{table*}

\begin{table*}[t!]
\centering
\small
\begin{tabularx}{\linewidth}{XXX}
\hline
\textbf{Input sentence} & \textbf{Sentence simplification} & \textbf{Conciseness model output} \\
\hline
A mutant is a type of fictional character that appears in comic books published by Marvel comics. & A mutant is a \textbf{form} of \textbf{imaginary} character that \textbf{is seen} in comic books published by Marvel comics. & A mutant \textbf{is a fictional} character that appears in \textbf{comics} published by Marvel comics. \\
\hline
It will then dislodge itself and sink back to the river bed in order to digest its food and wait for its next meal. & It will then \textbf{get away from its place} and sink back \textbf{into} the river bed in order to digest its food and wait for its next meal. & It will then dislodge and \textbf{return} to the riverbed to digest its food and wait for the next meal. \\
\hline
\end{tabularx}
\caption{\label{tab:examples-simpl-conciseness} Example outputs of one of our conciseness models on sentences from a text simplification data set \citep[WikiLarge]{zhang-lapata-2017-sentence}.}
\end{table*}

We evaluate a range of models on our newly collected conciseness test sets. Our initial approach follows the recent paradigm of using massively pre-trained neural models with either no or very little task-specific training data. Inspired by \citet{gpt-few-shot} we report on zero-shot experiments with the large language model LaMDA \citep{lamda}. We also fine-tune the large sequence model T5 \citep{t5} on small conciseness data sets.
We achieve our best results using an unsupervised synthetic data generation method based on round-trip translations, i.e. sentence pairs that were generated by translating an English sentence into another language (e.g.\ German) and back, a technique that was previously proposed for GEC pre-training~\citep{lichtarge-etal-2019-corpora}. We construct additional data sets by creating mappings from the longest to the shortest reference in multi-reference machine translation (MT) test sets. Our experiments suggest that conciseness is a hard task for current NLP models. We conclude with a thorough investigation into the similarities and differences of our systems and map out the challenges ahead.

\section{The conciseness task}
\label{sec:conciseness-task}

In this work we define the conciseness task as {\em applying the required edits to make a sentence less wordy without changing its meaning, intent or sentiment}. We will shed more light on the limitations of this definition in Sec.\ \ref{sec:limitations}. We expect conciseness models to be useful mainly for native or advanced non-native writers who wish to improve their writing style. Conciseness is related to several other NLP tasks, but we argue below that each of these tasks has a different focus and deserves an independent treatment.

\paragraph{Summarization and sentence compression}

Abstractive sentence summarization \citep{over-duc-summarization} attempts to produce a condensed version of the input text. Summaries are similar to headlines with a maximum length that is independent of the input sentence length \citep{rush-etal-2015-neural}. Thus, generating a summary often requires a much more severe compression compared to conciseness. Unlike summarization, conciseness is faithful to the input and aims to avoid the loss of any information -- the goal is to generate a shorter sentence that can replace the original sentence within continuous text (see Table \ref{tab:examples-summ-conciseness} for examples). Furthermore, most work on summarization focuses on the compression of entire documents or paragraphs \citep{pegasus} and not on single sentences.

Similarly to sentence summarization, {\em sentence compression} also aims to generate a shorter version of the input text. Many sentence compression models only allow the deletion of words without the ability to rephrase parts of the sentence \citep{knight-compress,jing-2000-sentence,filippova-etal-2015-sentence}. Perhaps closest to our work, \citet{mallinson-etal-2018-sentence} trained sentence compression models on round-trip translations and thereby avoided this restriction. The main difference to us is that we evaluate a broader range of methods on human-annotated test sets which we release for future research.

\paragraph{Sentence simplification}

The task of reducing the linguistic complexity of text to improve readability is known as {\em sentence simplification} \citep{simplification}. It can be subdivided into lexical (e.g.\ replacing uncommon words with synonyms) and syntactic (e.g.\ changing passive to active) simplification \citep{simplifying-phd,carroll-etal-1999-simplifying}. Most forms of syntactic simplification result in concise outputs,\footnote{An exception would be {\em  sentence splitting} since it is a syntactic simplification strategy that often makes the text longer.} but lexical simplification may yield even more verbose outputs. For example, replacing `to portray' with a simpler but verbose phrase such as `to describe very vividly' would be an instance of lexical simplification but not of conciseness. Conversely, a conciseness system may substitute a phrase with another that is concise but less common and thereby deteriorate readability. Another difference is that simplification often targets people with cognitive disabilities \citep{simplifying-phd,carroll-etal-1999-simplifying,simpl-dyslexia} or low literacy \citep{simpl-low-lit} or second language learners \citep{simpl-corpus-ana,siddharthan-simplification,xia-etal-2016-text} whereas conciseness can be thought as writing assistance for proficient writers. Table \ref{tab:examples-simpl-conciseness} contrasts simplification and conciseness with the help of example sentences.

\paragraph{Style transfer}

Text style is an important consideration for several NLP tasks \citep{style-transfer-exp-eval}. For example, it is desirable for MT output to match the stylistic properties of the source sentence \citep{sennrich-etal-2016-controlling,mt-maintaining-sentiment}. Natural language generation systems not only  need to take into account the content of generated utterances but also other attributes such as style and sentiment \citep{li-etal-2018-delete}. Text-to-text style transfer systems have been used to change Shakespearean English to modern English \citep{jhamtani-etal-2017-shakespearizing}. We consider conciseness as a special case of style transfer with a single source style (wordy) and one target style (concise). However, while most style transfer systems attempt to change attributes like sentiment or political slant \citep{li-etal-2018-delete,style-transfer-exp-eval,prabhumoye-etal-2018-style,style-transfer-shen}, our conciseness models aim to keep them unchanged.

\paragraph{Paraphrasing}

Paraphrasing databases such as PPDB \citep{ganitkevitch-etal-2013-ppdb,pavlick-etal-2015-ppdb} that store pairs of phrases with the same meaning have proven useful for various NLP tasks such as textual entailment \citep{bjerva-etal-2014-meaning} and semantic similarity \citep{han-etal-2013-umbc}. In this work we include a paraphrasing system for comparison.

\begin{figure*}[t!]
\centering
\small
\includegraphics[width=1.0\textwidth]{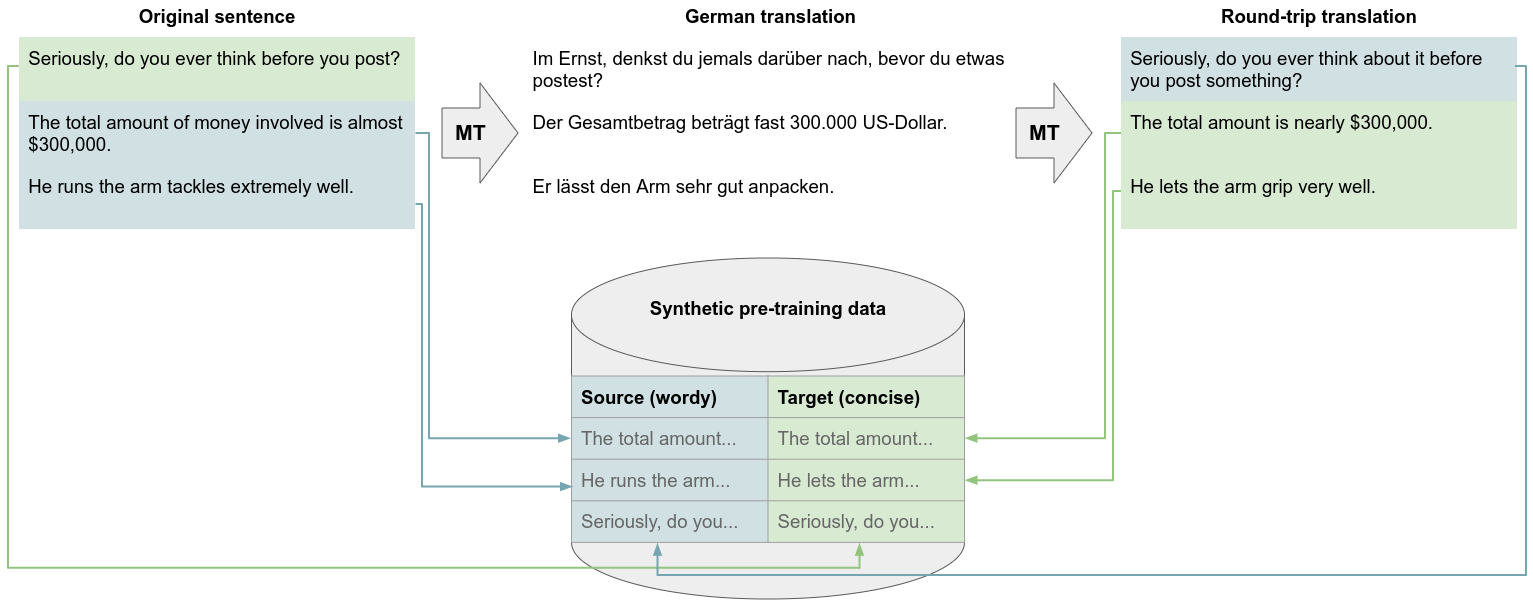}
\caption{Synthetic pre-training data generation using round-trip translations.}
\label{fig:rt}
\end{figure*}

\section{Modeling conciseness}
\label{sec:modeling-conciseness}

The approaches in this section cover a wide range of NLP models to convey a better sense for the task. They are intended to serve as baselines to compare against, and as a starting point for future research.

\subsection{Giant language models (LaMDA)}

Large language models (LMs) such as OpenAI's GPT-3 \citep{gpt}, Google's Meena \citep{meena} and PaLM \citep{palm2022} and Microsoft's Turing NLG\footnote{\url{https://msturing.org/}} have recently captured the interest of the general public through their ability to generate text that is sometimes astonishingly difficult to distinguish from text written by humans. While these models are useful for building open-domain dialog agents, they also have the potential to solve specific NLP problems when provided with an appropriate preamble (LM history) \citep{gpt-few-shot}. We expect general dialog agents to understand the nuances of language such as grammar, conciseness, etc. Thus, we explored using the large LM LaMDA \citep{lamda} with a zero-shot preamble that steers the model towards making a sentence more concise. We use the following template to provide the LM context:
\begin{quote}
\raggedright{
{\em Here is some text:  {``\texttt{[INPUT\_SENTENCE]}''}. Rewrite it to be more concise.}}
\end{quote}
where \texttt{[INPUT\_SENTENCE]} is replaced by the source sentence.\footnote{This prompt was best among a small number of zero-shot and few-shot prompts we explored. Systematic prompt engineering could potentially improve LaMDA results at a significantly higher computational cost, but we have not explored this option in this work since we focus on conciseness as an NLP task.} 
We post-process the output to a) discard any additional comment that the model generated besides the rewrite, and b) retain only the first suggestion if multiple rewrites are generated.

\begin{table*}
\centering
\small
\begin{tabular}{lcccc}
\hline \textbf{Name} & \textbf{Number of} & \textbf{Average source sentence} & \textbf{Average target sentence} & \textbf{Compression} \\
 & \textbf{sentence pairs} & \textbf{length in words} & \textbf{length in words} & \textbf{ratio} \\
\hline
\multicolumn{5}{l}{\textbf{Pre-training and fine-tuning data sets}} \\
\hline
RoundTrip-French & 169M & 20.6 & 19.4 & 0.94 \\
RoundTrip-German & 169M & 20.4 & 19.4 & 0.95 \\
RoundTrip-Japanese & 169M & 20.4 & 17.9 & 0.88 \\
RoundTrip-Russian & 169M & 20.9 & 19.5 & 0.93 \\
MultiRefMT-FineTune & 9K & 31.9 & 26.1 & 0.82 \\
\hline
\multicolumn{5}{l}{\textbf{Development sets}} \\
\hline
MultiRefMT-Dev & 820 & 33.3 & 25.8 & 0.77 \\
\hline
\multicolumn{5}{l}{\textbf{Hand-annotated test sets}} \\
\hline
Concise-Lite & 2K & 23.7 & 21.2 & 0.89 \\
Concise-Full & 2K & 23.7 & 20.1 & 0.85 \\
\hline
\end{tabular}
\caption{\label{tab:data-sets-stats} Data set statistics. The compression ratio is the number of target words divided by the number of source words.}
\end{table*}

\begin{table*}
\centering
\small
\begin{tabular}{lll}
\hline \textbf{Name} & \textbf{Reference} & \textbf{Type} \\
\hline
RoundTrip-* & \citet{lichtarge-etal-2019-corpora} & Round-trip translations (news) \\
MultiRefMT-FineTune & \href{https://catalog.ldc.upenn.edu/LDC2010T10}{LDC2010T10}, \href{https://catalog.ldc.upenn.edu/LDC2010T11}{LDC2010T11}, & 4-annotator MT test sets (Arabic-English, \\
 & \href{https://catalog.ldc.upenn.edu/LDC2010T12}{LDC2010T12}, \href{https://catalog.ldc.upenn.edu/LDC2010T14}{LDC2010T14} & Chinese-English) \\
MultiRefMT-Dev & \href{https://catalog.ldc.upenn.edu/LDC2013T03}{LDC2013T03} & 4-annotator MT test set (Chinese-English) \\
Concise-Lite & This work & 2-way hand-annotated conciseness test set \\
Concise-Full & This work & 5-way hand-annotated conciseness test set \\
\hline
\end{tabular}
\caption{\label{tab:data-sets} Synthetic and hand-annotated conciseness data sets used in this work.}
\end{table*}

\subsection{Transformers pre-trained on round-trip translations}
\label{sec:rt}

This method employs synthetic training data generated using MT. Fig.\ \ref{fig:rt} illustrates the approach. First, we translate an English sentence into a pivot language such as German, and then translate it back into English. This idea of generating sentence pairs via round-trip translation was initially proposed by \citet{lichtarge-etal-2019-corpora} to pre-train GEC systems. In this work, we construct synthetic parallel data for conciseness by using the longer sentence as the source and the shorter sentence as the target sentence. We then train a standard neural sequence-to-sequence Transformer \citep{transformer} on the synthetic data until convergence.\footnote{More details about the Transformer model implementation are provided in Appendix \ref{sec:trans-hyper}.}
This approach is simple and enables us to generate large quantities of data, but the resulting data set contains noise. For example, round-trip translation pairs often contain synonym substitutions (see the replacement of {\em almost} with {\em nearly} in the second sentence in Fig.\ \ref{fig:rt}) that do not help conciseness. Furthermore, MT may fail to translate the sentence properly, resulting in an undesirable change of meaning (see the third sentence in Fig.\ \ref{fig:rt}). Another problem is that it is hard to control the compression ratio in the data set. Despite these limitations we show in Sec.\ \ref{sec:results} that round-trip translations are useful for pre-training.

\begin{figure*}[t!]
\centering
\small
\includegraphics[width=1.0\textwidth]{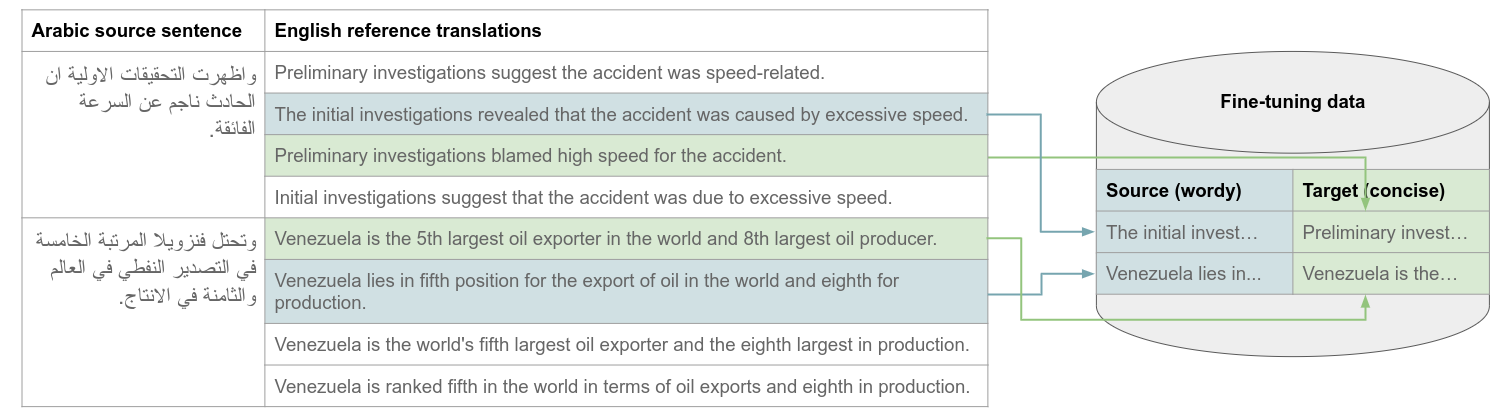}
\caption{Fine-tuning data generation using multi-reference MT test sets.}
\label{fig:ft}
\end{figure*}

\subsection{Fine-tuning T5}
\label{sec:t5}

The final method considered in this work employs T5 \citep{t5}. Very large sequence-to-sequence models have been found to be extremely powerful, even for challenging language tasks with a limited amount of training data. We fine-tuned the publicly available 11B parameter version (xxl) of T5\footnote{\url{https://github.com/google-research/text-to-text-transfer-transformer/blob/main/released_checkpoints.md}}, with a batch size of 1,024 sentences and a learning rate of $10^{-4}$. 

\section{Data sets}

Table \ref{tab:data-sets-stats} lists the data sets used in this work. Table \ref{tab:data-sets} contains information about their provenance.

\paragraph{Round-trip translations (RoundTrip-*)}

Our Transformer system is pre-trained on round-trip translations of sentences crawled from news websites following the recipe of \citet{lichtarge-etal-2019-corpora} that were prepared as described in Sec.\ \ref{sec:rt}. For fine-tuning T5 on round-trip translations we randomly sample 1M sentence pairs from the full data set to limit computation.

\paragraph{OpenMT-based fine-tuning and development sets (MultiRefMT-*)}

We derive fine-tuning and development sets from existing publicly available MT test sets. It is common practice in several NLP areas to collect reference sentences from multiple annotators to increase the trustworthiness of automatic evaluation measures, for example in grammatical error correction \citep{ng-etal-2014-conll,bryant-ng-2015-far,napoles-etal-2017-jfleg}, MT \citep{freitag-etal-2020-bleu}, and image caption generation  \citep{zheng-etal-2018-multi}. Multi-reference MT test sets have been used in the past to evaluate paraphrasing or sentence compression systems \citep{ganitkevitch-etal-2011-learning,pang-etal-2003-syntax}. We make use of these multi-annotator test sets by selecting the longest reference sentence as the (wordy) source sentence and the shortest reference sentence as the golden (concise) target sentence (Fig.\ \ref{fig:ft}). Our MultiRefMT-FineTune set uses all Arabic-English and Chinese-English NIST Open Machine Translation (OpenMT) evaluation sets from 2002-2005. The MultiRefMT-Dev set is based on the Chinese-English 2012 OpenMT evaluation set.

\paragraph{Hand-annotated test sets (Concise-*)}

Deriving conciseness test sets from multi-reference MT evaluation sets is viable as a first approximation given that all references have similar meaning, intent, and sentiment by design (apart from annotation errors). However, it does not allow us to determine how wordy the sentence is in the first place. If all MT references agreed, it would suggest that the original source sentence has a single obvious translation, not that the references are already concise.

Therefore, we collected two new data sets, consisting of 2000 sentences each, that were explicitly annotated for conciseness -- {\em Concise-Lite} and {\em Concise-Full}. Both data sets used the same set of source sentences drawn from Wikipedia. Sentences that a) were ungrammatical, b) contained fewer than 15 words or c) included mismatched quotation marks were not selected. While {\em Concise-Lite} annotators were asked to make minimal changes to the original sentence, {\em Concise-Full} annotators were given the flexibility to make larger changes to the original sentence. The exact annotator guidelines are listed in Appendix \ref{sec:annotator-instructions}.

We will make the test sets publicly available to establish a benchmark for researchers to evaluate conciseness models.

\begin{table*}
\centering
\small
\begin{tabular}{@{\hspace{0em}}l@{\hspace{0.3em}}lcccccc}
\hline
& \textbf{System} &  \multicolumn{3}{c}{\textbf{Concise-Lite}} & \multicolumn{3}{c}{\textbf{Concise-Full}} \\
 &  & P & R & \cellcolor{f05} $F_{0.5}$  & P & R & \cellcolor{f05} $F_{0.5}$   \\
\hline
\multicolumn{8}{l}{\textbf{Other NLP tasks}} \\
\hline
\footnotesize{a} & Summarization: Pegasus & \ 0.8 & \ 1.4 & \cellcolor{f05} \ 0.9 & \ 2.0 & \ 3.9 & \cellcolor{f05} \ 2.2  \\
\footnotesize{b} & Summarization: Long-T5 & \ 1.7 & \ 6.3 & \cellcolor{f05} \ 2.0 & \ 3.5 & 11.7 & \cellcolor{f05} \ 4.1  \\
\footnotesize{c} & Simplification: T5 & \ 7.4 & \ 5.4 & \cellcolor{f05} \ 6.9 & 13.8 & \ 9.9 & \cellcolor{f05} 12.8  \\
\footnotesize{d} & Paraphrasing: ParaNMT & \ 9.3 & 21.4 & \cellcolor{f05} 10.4 & 15.4 & 25.1 & \cellcolor{f05} 16.7  \\
\hline
\multicolumn{8}{l}{\textbf{Conciseness models}} \\
\hline
\footnotesize{e} & Giant-LM (zero-shot LaMDA) & \ 4.4 & 13.5 & \cellcolor{f05} \ 5.1 & \ 8.5 & 20.0 & \cellcolor{f05} \ 9.6  \\
\footnotesize{f} & Transformer (RT) & 13.6 & 21.3 & \cellcolor{f05} 14.6 & 21.1 & 25.5 & \cellcolor{f05} 21.9  \\
\footnotesize{g} & Transformer (RT$\rightarrow$MT) & 15.0 & 25.8 & \cellcolor{f05} 16.4 & 24.4 & 29.6 & \cellcolor{f05} 25.2  \\
\footnotesize{h} & T5 (RT)   & 18.4 & 19.5 & \cellcolor{f05} 18.6 & 29.1 & 24.2 & \cellcolor{f05} 28.0 \\
\footnotesize{i} & T5 (RT$\rightarrow$MT) & 16.0 & 26.8 & \cellcolor{f05} 17.4 & 26.6 & 30.6 & \cellcolor{f05} 27.3 \\
\hline
\end{tabular}
\caption{\label{tab:system-results} System comparison on our two conciseness test sets. ``RT'' denotes models trained on round-trip translations. ``RT$\rightarrow$MT'' configurations are subsequently fine-tuned on MultiRefMT-FineTune.}
\end{table*}

\begin{table}[t!]
\centering
\small
\begin{tabular}{ll}
\hline \textbf{System} & \textbf{Number of parameters} \\ \hline
Giant-LM (LaMDA) & 137B \\
T5 & 11B \\
Transformer & 313M \\
\hline
\end{tabular}
\caption{\label{tab:params} Number of model parameters.}
\end{table}

\section{Results}
\label{sec:results}

We use the GEC evaluation toolkit ERRANT \citep{bryant-etal-2017-automatic,felice-etal-2016-automatic} to compute $F_{0.5}$-scores on spaCy\footnote{\url{https://spacy.io/}}-tokenized text. Like in GEC, precision is weighted twice as high as recall using the $F_{0.5}$-score, which matches our intuition that a conciseness system should act as a minimally intrusive writing assistant for which false positives are far worse than false negatives.

\subsection{System comparison}

\begin{figure}[t!]
\centering
\small
\includegraphics[width=0.28\textwidth]{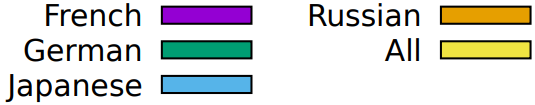}

\vspace{-2.0em}

\includegraphics[scale=1.0]{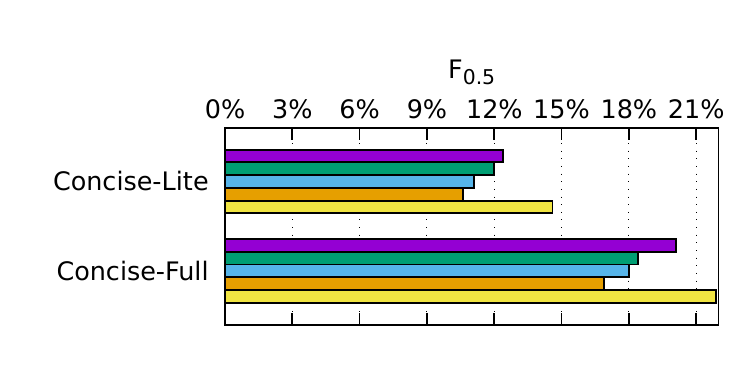}
\caption{Transformer models trained from scratch on round-trip translations via different pivot languages.}
\label{fig:rt-langs}
\end{figure}

Table \ref{tab:system-results} compares all approaches from Sec.\ \ref{sec:modeling-conciseness} and the following baselines from other NLP tasks:
\begin{itemize}
    \item Summarization: Long-T5 \citep{longt5} and Pegasus \citep{pegasus}.
    \item Simplification: T5 fine-tuned on the WikiLarge simplification dataset \citep{zhang-lapata-2017-sentence} using a procedure similar to our T5-conciseness system from Sec.\ \ref{sec:t5}.\footnote{Our simplification baseline achieves 33.1 SARI on the WikiLarge test set.}
    \item Paraphrasing: A Transformer model trained on the full ParaNMT-50M \citep{wieting-gimpel-2018-paranmt} training set using the hyper-parameters in Appendix \ref{sec:trans-hyper}.
\end{itemize}

The summarization baselines (rows a and b) perform poorly since they are mostly trained on full documents. 
The simplification system achieves a slightly higher performance but is weaker than the paraphrasing or the Transformer/T5 based conciseness systems. The paraphrasing system (row d) achieved a recall of over 20\% on both test sets, but the precision is relatively low because the ParaNMT training set contains various types of edits such as synonym replacements or word reorderings that do not necessarily help conciseness. 

The zero shot Giant-LM (LaMDA) setup (row e) was not able to match either the precision or recall of the other conciseness systems.  Round-trip translations are useful for both training a Transformer model from scratch (row f) and fine-tuning T5 (row h). Subsequent fine-tuning on MultiRefMT-FineTune yields large precision and recall gains for the Transformer model (row g). MultiRefMT-FineTune also improves the recall for T5, but the precision suffers (row i).\footnote{T5 is fine-tuned for 4K steps on the 1M round-trip translations and for 1K steps on the smaller MultiRefMT-FineTune set.} T5 outperforms the Transformers in terms of $F_{0.5}$-score by achieving higher precision on both sets but has many more parameters (Table \ref{tab:params}).

\subsection{Ablation studies and analyses}
\label{sec:ablation}

The following analyses were carried out on the {\em Concise-Lite} and {\em Concise-Full} test sets.

\paragraph{Round-trip translation languages}

Our final models in Table \ref{tab:system-results} use round-trip translations from four different pivot languages: French, German, Japanese, and Russian. Fig.\ \ref{fig:rt-langs} shows that combining all languages yields consistent gains on both test sets over using any single language.

\begin{figure}[t!]
\centering
\small
\includegraphics[scale=1.0]{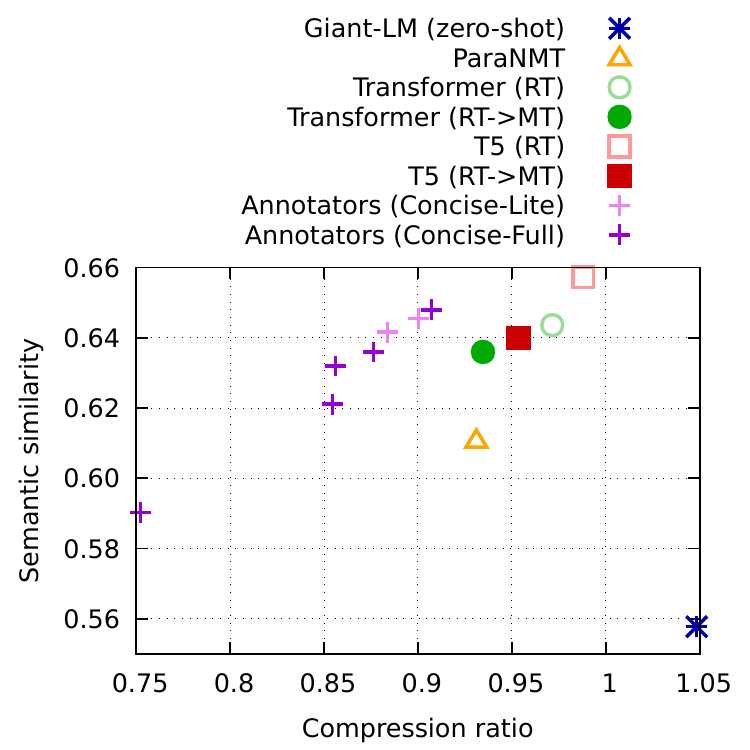}
\caption{Trade-off between semantic similarity and the sentence compression ratio.}
\label{fig:cr_kona}
\end{figure}

\paragraph{Preserving semantics}

To measure how well our systems retain the meaning of the original sentence we computed semantic similarity scores between the input and the output sentences using the models provided by the Semantic Reactor toolkit \citep{yang-etal-2018-learning,uni-sen-encode}. Systems and annotators trade off compression against semantic similarity differently (Figure~\ref{fig:cr_kona}). There is a large variability in compression ratio (i.e.\ the number of target words divided by the number of source words) and semantic similarity between the {\em Concise-Full} annotators (dark purple). The Giant-LM (blue) is more prone to meaning change than other systems, and is not effective in reducing the sentence length. Fine-tuning on MultiRefMT-FineTune (empty vs.\ filled circle/square) improves the compression ratio but hurts semantic similarity. T5 (red) preserves semantics better than the Transformer but outputs slightly longer sentences.

\paragraph{Readability}

\begin{figure}[t!]
\centering
\small
\includegraphics[scale=1.0]{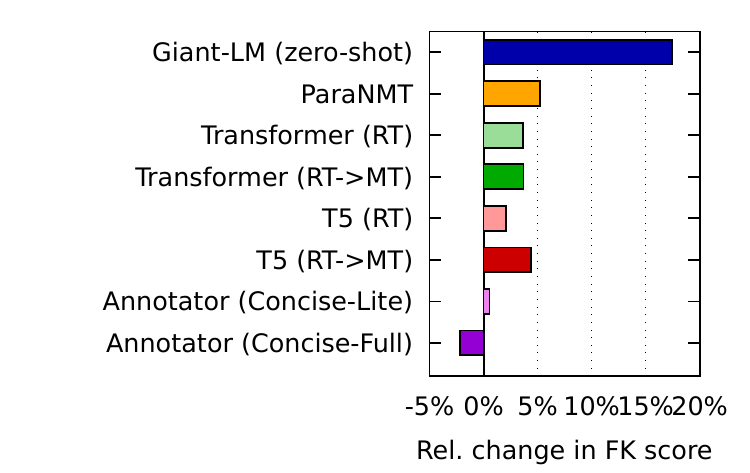}
\caption{Relative change in Flesch–Kincaid readability scores \citep{fk}.}
\label{fig:fk}
\end{figure}

Fig.\ \ref{fig:fk} shows that our systems often improve the readability of the sentence, in particular the Giant-LM system. The Giant-LM prefers simpler language as it was originally designed for dialog applications \citep{lamda}. In contrast, the {\em Concise-Full} annotators tend to achieve concision using longer and more complex words, resulting in a decline in readability (dark purple).

\begin{figure}[t!]
\centering
\small
\includegraphics[scale=1.0]{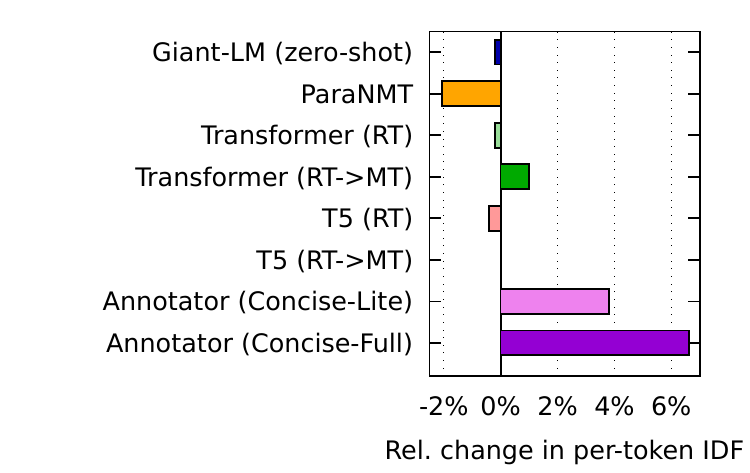}
\caption{Relative change in information density.}
\label{fig:idf}
\end{figure}

\paragraph{Information density}
We expect the outputs of a high-performing conciseness system to have a high information content per word.  This information density can be measured using per-token inverse document frequency~\citep{jones1973idf}: $$\text{idf}(t) = \log \: \frac{N}{|\{d \in D : t \in d\}|},$$
where $t$ is the token, $N$ is the total number of documents, and $D$ is the document collection. In our case, the document frequencies are derived from the C4 corpus~\citep{t5}. Fig.\ \ref{fig:idf} shows that the reference sentences from the {\em Concise-Lite} and {\em Concise-Full} annotators indeed have a higher per-token IDF than the input sentences (pink and dark purple bars). The results on the system outputs are mixed, but fine-tuning on MultiRefMT-FineTune improves the per-token IDF for the Transformer and T5 (``RT'' vs.\ ``RT$\rightarrow$ MT'').

\begin{figure}[t!]
\centering
\small
\includegraphics[scale=1.0]{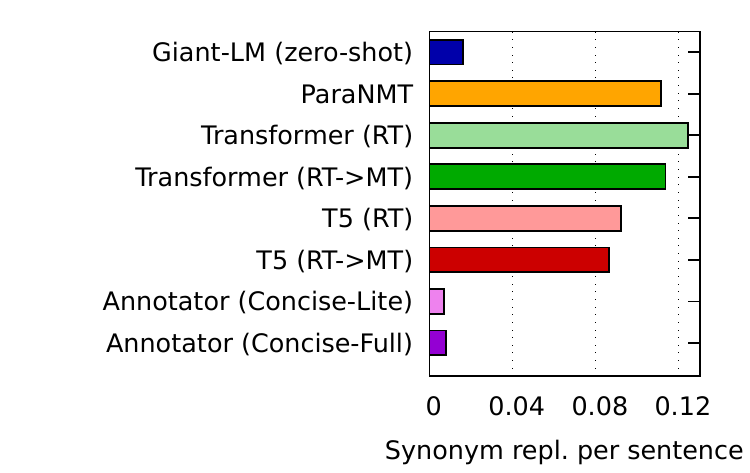}
\caption{Number of 1:1 synonym substitutions.}
\label{fig:syn}
\end{figure}

\begin{table*}
\centering
\small
\begin{tabular}{l|c@{\hspace{0.9em}}c@{\hspace{0.9em}}c@{\hspace{0.9em}}c@{\hspace{0.9em}}c@{\hspace{0.9em}}c@{\hspace{0.9em}}c@{\hspace{0.9em}}c@{\hspace{0.9em}}c@{\hspace{0.9em}}c@{\hspace{0.9em}}c@{\hspace{0.9em}}c@{\hspace{0.9em}}c@{\hspace{0.9em}}c@{\hspace{0.9em}}c}
\hline
 &  \multicolumn{3}{c}{\textbf{Without A1}} &  \multicolumn{3}{c}{\textbf{Without A2}} &  \multicolumn{3}{c}{\textbf{Without A3}} &  \multicolumn{3}{c}{\textbf{Without A4}} &  \multicolumn{3}{c}{\textbf{Without A5}}  \\
  & P & R & \cellcolor{f05} $F_{0.5}$  & P & R & \cellcolor{f05} $F_{0.5}$ & P & R & \cellcolor{f05} $F_{0.5}$ & P & R & \cellcolor{f05} $F_{0.5}$ & P & R & \cellcolor{f05} $F_{0.5}$  \\
\hline
Annotator A1 & 45.8 & 52.0 & \cellcolor{f05} 46.9 &  & & \cellcolor{f05} &  & & \cellcolor{f05} &  & & \cellcolor{f05} &  & & \cellcolor{f05} \\
Annotator A2 &  & & \cellcolor{f05} & 16.3 &  32.0 & \cellcolor{f05} 18.1 &  & & \cellcolor{f05} &  & & \cellcolor{f05} &  & & \cellcolor{f05} \\
Annotator A3 &  & & \cellcolor{f05} &  & & \cellcolor{f05} & 51.5 & 48.4 & \cellcolor{f05} 50.9 &  & & \cellcolor{f05} &  & & \cellcolor{f05} \\
Annotator A4 &  & & \cellcolor{f05} &  & & \cellcolor{f05} &  & & \cellcolor{f05} & 23.1 & 32.6 & \cellcolor{f05} 24.5 &  & & \cellcolor{f05} \\
Annotator A5 &  & & \cellcolor{f05} &  & & \cellcolor{f05} &  & & \cellcolor{f05} &  & & \cellcolor{f05} & 33.5 & 27.1 & \cellcolor{f05} 32.0 \\
\hline
Transformer & 22.7  & 27.6 & \cellcolor{f05} 23.5 & 19.7 & 28.6 & \cellcolor{f05} 21.0 & 23.6 & 27.7 & \cellcolor{f05} 24.3 & 20.8  & 26.8 & \cellcolor{f05} 21.8 & 23.0  & 25.9 & \cellcolor{f05} 23.6 \\
T5 & 25.3 & 28.9 & \cellcolor{f05} 26.0 & 20.7  & 29.2 & \cellcolor{f05} 22.0 & 25.7 & 28.9 & \cellcolor{f05} 26.3 & 23.1 & 28.0 & \cellcolor{f05} 23.9 & 25.4  & 27.0 & \cellcolor{f05} 25.7 \\
\hline
\end{tabular}
\caption{\label{tab:leave-one-out} Measuring annotator agreement on {\em Concise-Full} by evaluating each single annotator using the other four annotations as references. We list the Transformer and T5 system outputs (``RT$\rightarrow$MT'') for comparison.}
\end{table*}

\paragraph{Synonym substitutions}

One problem with using round-trip translations for training and multi-reference test sets for evaluation is that both may contain synonym substitutions that do not help conciseness. We counted synonym substitutions by extracting all 1:1 substitutions and checking whether these were marked as synonyms in WordNet \citep{wordnet}. Fig.\ \ref{fig:syn} shows that most of our systems replace synonyms on an average in every 10th sentence. Fine-tuning the Transformer or T5 on MultiRefMT-FineTune reduces the number of synonym substitutions. Synonyms are much less of a problem with the Giant-LM (blue bar) which was not trained on round-trip translations.

\section{Limitations}
\label{sec:limitations}

In terms of both information density (Fig.\ \ref{fig:idf}) and number of unnecessary synonym replacements (Fig.\ \ref{fig:syn}), the annotators are clearly separated from most of our automatic systems, illustrating the gap to human performance on this task.

Our experiments showed that the Giant-LM (zero-shot) underperformed the other approaches. Preliminary experiments using few-shot learning did not yield improvements over the zero-shot setting. We expect the performance of Giant-LM to improve via systematic prompt engineering.

Another challenge lies in the intrinsic uncertainty \citep{nmt-uncertainty,stahlberg-etal-2022-uncertainty} of the conciseness task, i.e.\ the existence of multiple viable ways to make a sentence more concise. Table \ref{tab:leave-one-out} demonstrates that the five {\em Concise-Full} annotators usually did not agree on a single concise version of a sentence, leading to great variability in $F_{0.5}$-scores when evaluated against each other.\footnote{On some of the setups in Table \ref{tab:leave-one-out} (e.g.\ ``Without A2'' or ``Without A4''), T5 achieves scores comparable to the human annotators. We emphasize that this is a sign of low inter-annotator agreement and does not allow us to claim human parity since this pattern is not consistent across annotators.} Therefore, adequate system outputs may get penalized if they do not agree with one of the human references. We mitigate this concern by using multiple annotators, but -- like in other intrinsically uncertain NLP tasks such as MT -- a certain level of noise remains in our evaluation.

\paragraph{Limitations of our task definition}

We acknowledge that there are various aspects of conciseness that are not covered by our definition in Sec.\ \ref{sec:conciseness-task} (``{\em applying the required edits to make a sentence less wordy without changing its meaning, intent or sentiment}''). First, we intentionally did not include the use of context in our definition. In practice, however, appropriate levels of conciseness can be highly context dependent. Treating the problem on the sentence-level is limiting because using inter-sentential cross-references for conciseness requires access to the document-level context such as the previous sentence. Furthermore, the sentence-level restriction prevents the systems from improving conciseness through sentence splitting \citep{botha-etal-2018-learning} or merging \citep{geva-etal-2019-discofuse}. In real-life situations, the context may also be provided through other channels such as physical medium (e.g.\ pointing to things) or social factors (e.g. does person B know person A?). We also noticed that our {\em Concise-Full} annotators occasionally relied on common knowledge to shorten sentences (see Appendix \ref{sec:example-outputs} for examples), a strategy that is {\em not} covered by our definition and thus makes our evaluation slightly more noisy. Exploring the various forms of context for conciseness is a promising potential direction for future research.

Another limitation of our definition is that it does not allow for a change of semantics, intent, or sentiment. In practice, however, conciseness or the lack of it may reflect the intent of the speaker, for example in indicating emergency situations (signalling urgency through brevity) or in detecting lying \citep{cbca}. Another manner in which conciseness can carry meaning is when used as a rhetorical device to persuade or inspire the audience, a well-known strategy in legal writing \citep{legal-writing} that was perhaps most famously demonstrated by Abraham Lincoln in the Gettysburg Address \citep{legal-brevity}. Furthermore, our ablation studies in Sec.\ \ref{sec:ablation} revealed that systems and human annotators alike sometimes accepted a minor loss of (irrelevant) information to achieve better compression, which, despite being contrary to our definition, may be acceptable in practice.

\section{Conclusion}

Our work is an initial exploration of conciseness from an NLP point of view. We compared a variety of approaches to the problem using popular techniques based on synthetic data generation or giant pre-trained sequence models. Round-trip translations provide a useful data source for training conciseness models but can introduce undesirable synonym substitutions.\footnote{Appendix \ref{sec:example-outputs} illustrates the strengths and weaknesses of our current systems with the help of some example outputs.} Our analyses show that our systems trade off the objectives in conciseness differently (e.g. reducing the sentence length vs.\ preserving semantics vs.\ improving readability vs.\ increasing information density). Further experiments are necessary to understand how these trade-offs would impact the user experience or potential downstream NLP tasks.
We expect our study and our annotated test sets to provide impetus for researchers to explore this field further.

\bibliography{anthology,custom}
\bibliographystyle{acl_natbib}

\clearpage

\appendix
\onecolumn

\section{Transformer hyper-parameters}
\label{sec:trans-hyper}

Our round-trip translation based models (Sec.\ \ref{sec:rt}) are trained on TPUs with the {\sc Lamb} optimizer \citep{lamb} in JAX \citep{jax}. We used the Transformer \citep{transformer} implementation from the MT example in Flax\footnote{\url{https://github.com/google/flax/tree/master/examples/wmt/}} with the 32K SentencePiece vocabulary \citep{kudo-richardson-2018-sentencepiece} from T5 \citep{t5}. Model hyper-parameters are listed in Table \ref{tab:trans-hyper}.

\begin{table}
\centering
\small
\begin{tabular}{ll}
\hline \textbf{Parameter} & \textbf{Value} \\ \hline
Attention dropout rate & 0.1 \\
Attention layer size & 1,024 \\
Batch size & 256 \\
Beam size & 10 \\
Dropout rate & 0.1 \\
Embedding size & 1,536 \\
Learning rate & 0.4 \\	
MLP dimension & 4,096 \\	
Number of attention heads & 4 \\
Number of layers & 6 \\
Number of fine-tuning iterations & 100-2,000 \\
& (early stopping) \\
Number of pre-training iterations & 100,000 \\
TPU topology & 4x4 \\
\hline
\end{tabular}
\caption{\label{tab:trans-hyper} Transformer hyper-parameters.}
\end{table}

\section{Annotator instructions}
\label{sec:annotator-instructions}

The {\em Concise-Lite} annotators received the following instructions:
\begin{quote}
    {\em Rewrite the sentence to make it more concise, without changing the sentence structure. By sentence structure, we mean the general order of words in the sentence should not change, some sub-phrases could be rewritten/replaced/deleted (3-5 words).  These should be relatively minor rewrites, such that you can replace a phrase with a shorter alternative without reorganizing the entire sentence. The sentences should be annotated in isolation without any assumptions on preceding or succeeding sentences.}
\end{quote}
The {\em Concise-Full} instructions are:
\begin{quote}
    {\em Rewrite the sentence to achieve maximum conciseness. These can be major rewrites that alter the sentence structure to make it as concise as possible. 
The annotator needs to make sure that the sentence stays the same semantically (meaning, intent \& sentiment) and there is no loss of any critical information. The sentences should be annotated in isolation without any assumptions on preceding or succeeding sentences.}
\end{quote}

\section{Example outputs}
\label{sec:example-outputs}

Table \ref{tab:example-outputs} shows some example outputs of our systems and the baselines.  The summarization (Long T5) system frequently changes the meaning of the source sentence. The simplification 
(Simplify T5) system performs slightly better but still changes the meaning in some instances (example c).
The T5 system is mostly faithful to the meaning of the source sentence. We observe occasional slight meaning shifts with the Transformer and ParaNMT systems (see e.g.\ examples b) and g)). The Giant-LM often changes or expands the information in the source sentence (e.g.\ examples b) and d), f)) or adds certain artefacts (e.g.\ ``{\em Here is a revision: `\dots' }'' in example a)) that stem from its main use case as a user-facing dialog agent. Being a paraphrasing system, ParaNMT often falls short of actually improving the conciseness (examples c) and f)), and often uses unnecessary synonyms. Synonym replacements can also be found sometimes in Transformer and T5 outputs (examples a) and c)), but not in Giant-LM and human-annotated sentences. The pre-trained models Giant-LM and T5 are sometimes able to compress sentences by relying on background knowledge, e.g.\ by replacing ``the Northern States'' with ``the Union'' or ``the North'' in example g).

\begin{table*}[t!]
\centering
\scriptsize
\begin{tabularx}{\linewidth}{lX}
\hline
\small \textbf{Example a)} & \\
\hline
Source & A few other men fell and sneaked back, but I don't know their names.\\
Long T5 &   A few other men fell and sneaked back. \\
Simplify T5 & A few other men fell and sneaked back, but I don't know their names. \\
Giant-LM & Here is a revision: Other men fell and snuck back, but their names I do not know.\\
ParaNMT & several other men fell and crept back, but I don't know their names. \\
Transformer & Some other men fell and sneaked, but I do not know their names. \\
T5 & A few other men fell and escaped, but I do not know their names. \\
Human (C.-Lite) & Some other men fell and sneaked back, but I don't know their names.  \\
Human (C.-Full) &  Others fell and snuck back, but I don't know their names. \\
\hline
\small \textbf{Example b)} & \\
\hline
Source & "Helplessly Hoping" is a 1969 song by the American folk rock group Crosby, Stills, and Nash. \\
Long T5 & Helplessly Hoping is a 1969 song by the American folk rock group Crosby, Stills and Nash. \\
Simplify T5 & Helplessly Hoping is a song by the American folk rock group Crosby, Stills \& Nash . \\
Giant-LM & "Helplessly Hoping" is a 1969 song by the American folk rock group Crosby, Stills, and Nash. The song was written for Stephen Stills' wife. \\
ParaNMT & "helpless Hoping" is a 1969 song from the American rock band Crosby, Stills and Nash. \\
Transformer & "Helpless Hoping" is a 1969 song by American folk rock group Crosby, Stills and Nash. \\
T5 & "Helplessly Hoping" is a 1969 song by the American folk rock group Crosby, Stills and Nash. \\
Human (C.-Lite) & "Helplessly Hoping" is a 1969 song by American folk rock group Crosby, Stills, and Nash. \\
Human (C.-Full) & Crosby, Stills, and Nash's "Helplessly Hoping" was released in 1969. \\
\hline
\small \textbf{Example c)} & \\
\hline
Source & The NLA Tower, Britain's 88th tallest tower, is an example of original 1970s architecture in the town. \\
Long T5 & The NLA Tower, Britain's 88th tallest tower, is an example of original 1970s architecture in the town. \\
Simplify T5 & The NLA Tower is an example of original 1970s architecture in the town . \\
Giant-LM & NLA Tower, Britain's 88th tallest tower, is a 1970s example of architecture in the town. \\
ParaNMT & the NLA Tower, the 88th highest British tower, is an example of the architecture of the 1970s in the city. \\
Transformer & The NLA Tower, Britain's 88th tallest tower, is an example of the city's original 1970s architecture.\\
T5 & The NLA Tower, Britain's 88th tallest building, is an example of the town's original 1970s architecture. \\
Human (C.-Lite) & The NLA Tower, Britain's 88th tallest tower, is an original 1970s architecture in town. \\
Human (C.-Full) &  Britain's 88th tallest tower, the NLA Tower, is an original 1970s architecture. \\
\hline
\small \textbf{Example d)} & \\
\hline
Source & There are some claims that even the kaiser had been amused by the incident. \\
Long T5 & Police say the kaiser used a special case kaiser to cover the snafu on a McDonald's breakfast. \\
Simplify T5 & There are some claims that even the kaiser had been amused by the incident. \\
Giant-LM & the kaiser was reportedly amused by the incident --  This is a pretty vague phrase, in fact. Was the kaiser laughing? Grinning? \\
ParaNMT & there are some claims that even the kaiser was amused by the incident. \\
Transformer & There are some claims that even Kaiser was amused by the incident. \\
T5 & Some claim that even the Kaiser was amused by the incident. \\
Human (C.-Lite) &  There are claims that even the Kaiser was amused by the incident. \\
Human (C.-Full) & Some claim that the incident amused even the kaiser. \\
\hline
\small \textbf{Example e)} & \\
\hline
Source & Approximately 50\% of viewers spend between one and three hours watching TV per day. \\
Long T5 & TV Viewership Up to 50\%. \\
Simplify T5 & Approximately 50\% of viewers spend between one and three hours watching TV per day. \\
Giant-LM & About half of viewers watch TV for 1 to 3 hours each day. \\
ParaNMT & approximately 50\% of viewers spend one to three hours watching TV every day. \\
Transformer & About 50\% of viewers spend between one and three hours watching TV a day. \\
T5 &  About 50\% of viewers watch TV for one to three hours a day. \\
Human (C.-Lite) & Approximately 50\% of viewers spend between one and three hours watching TV daily.  \\
Human (C.-Full) & 50\% of viewers spend 1-3 hours watching TV daily. \\
\hline
\small \textbf{Example f)} & \\
\hline
Source & He studied at Tübingen, Freiburg, Berlin, and Heidelberg, where he earned his medical degree in 1910. \\
Long T5 & German Doctor Emil Henckel von Donnersmarck Dies at 87 \\
Simplify T5 & He studied at Tübingen, Freiburg, Berlin, and Heidelberg, where he earned his medical degree in 1910. \\ 
Giant-LM & He studied medicine in Tübingen, Freiburg, Berlin and Heidelberg. \\
ParaNMT & he studied at Tübingen, Freiburg, Berlin, and Heidelberg, where he earned a medical degree in 1910. \\
Transformer & He studied in Tübingen, Freiburg, Berlin and Heidelberg, where he graduated in medicine in 1910.\\
T5 & He studied in Tübingen, Freiburg, Berlin and Heidelberg, where he received his medical degree in 1910. \\
Human (C.-Lite) & He studied at Tübingen, Freiburg, Berlin, and Heidelberg, where he earned his medical degree in 1910. \\
Human (C.-Full) & He studied at Tübingen, Freiburg, Berlin, and Heidelberg, earning his medical degree in 1910. \\
\hline
\small \textbf{Example g)} & \\
\hline
Source & Almost without exception, the Keetoowahs sided with the Northern States during the Civil War. \\
Long T5 & Keetoowahs Speak Out on Civil War. \\
Simplify T5 & Almost without exception, the Keetoowahs sided with the Northern States during the Civil War. \\
Giant-LM & Almost without exception, the Keetoowahs sided with the Union.\\
ParaNMT &almost without exception, Keetoowah sailed with the Northern States during the Civil War.\\
Transformer & Almost without exception, the Keetoowahs joined the northern states during the civil war.\\
T5 & Almost without exception, the Keetoowahs sided with the North during the Civil War. \\
Human (C.-Lite) & The Keetoowahs sided with the Northern States during the Civil War. \\
Human (C.-Full) & During the Civil War, the Keetoowahs sided with the North. \\
\hline
\end{tabularx}
\caption{\label{tab:example-outputs} Example sentences from our conciseness systems and other baselines (summarization: Long T5, simplification: Simplify T5, ParaNMT). We use the ``RT$\rightarrow$ MT'' setups for the Transformer and T5 systems. We show one {\em Concise-Lite} and one {\em Concise-Full} human reference.}
\end{table*}

\end{document}